\documentclass[10pt,twocolumn,letterpaper]{article}

\usepackage{cvpr}              

\usepackage{graphicx}
\usepackage{amsmath}
\usepackage{amssymb}
\usepackage{booktabs}
\usepackage{xcolor}
\usepackage{MnSymbol}
\usepackage{array}
\usepackage[accsupp]{axessibility}
\newcolumntype{?}{!{\vrule width 1pt}}

\usepackage[pagebackref,breaklinks,colorlinks]{hyperref}

\usepackage[capitalize]{cleveref}
\crefname{section}{Sec.}{Secs.}
\Crefname{section}{Section}{Sections}
\Crefname{table}{Table}{Tables}
\crefname{table}{Tab.}{Tabs.}

\def\cvprPaperID{9027} 
\def\confName{CVPR}
\def\confYear{2022}

\begin{document}

\title{Towards Low-Cost and Efficient Malaria Detection}

\author{Waqas Sultani\textsuperscript{1}, Wajahat Nawaz\textsuperscript{1}\thanks{Authors contributed equally} , Syed Javed\textsuperscript{1*}, Muhammad Sohail Danish\textsuperscript{1*}, Asma Saadia\textsuperscript{2},
Mohsen Ali\textsuperscript{1}\\
\textsuperscript{1}Intelligent Machines Lab, Information Technology University, Pakistan\\
\textsuperscript{2}Central Park Medical College, Pakistan\\
}
\maketitle
\begin{abstract}
Malaria, a fatal but curable disease claims hundreds of thousands of lives every year.
Early and correct diagnosis is vital to avoid health complexities, however, it depends upon the availability of costly microscopes and trained experts to analyze blood-smear slides. 
Deep learning-based methods have the potential to not only decrease the burden of experts but also improve diagnostic accuracy on low-cost microscopes. 
However, this is hampered by the absence of a reasonable size dataset.
One of the most challenging aspects is the reluctance of the experts to annotate the dataset at low magnification on low-cost microscopes.
We present a dataset to further the research on malaria microscopy over  low-cost microscopes at low magnification. 
Our large-scale dataset consists of images of blood-smear slides from several malaria-infected patients, collected through microscopes at two different cost spectrums and multiple magnifications. 
Malarial cells are annotated for the localization and life-stage classification task on the images collected through the high-cost microscope at high magnification.
We design a mechanism to transfer these annotations from the high-cost microscope at high magnification to the low-cost microscope, at multiple magnifications. 
Multiple object detectors and domain adaptation methods are presented as the baselines.  Furthermore, a partially supervised domain adaptation method is introduced to adapt the object-detector to work on the images collected from the low-cost microscope.  
The dataset is available here: \href{http://im.itu.edu.pk/m5-malaria-dataset/}{http://im.itu.edu.pk/m5-malaria-dataset/}


\end{abstract}

\section{Introduction}
\label{sec:intro}
\begin{figure}[t]
    \centering
    \includegraphics[width=\linewidth]{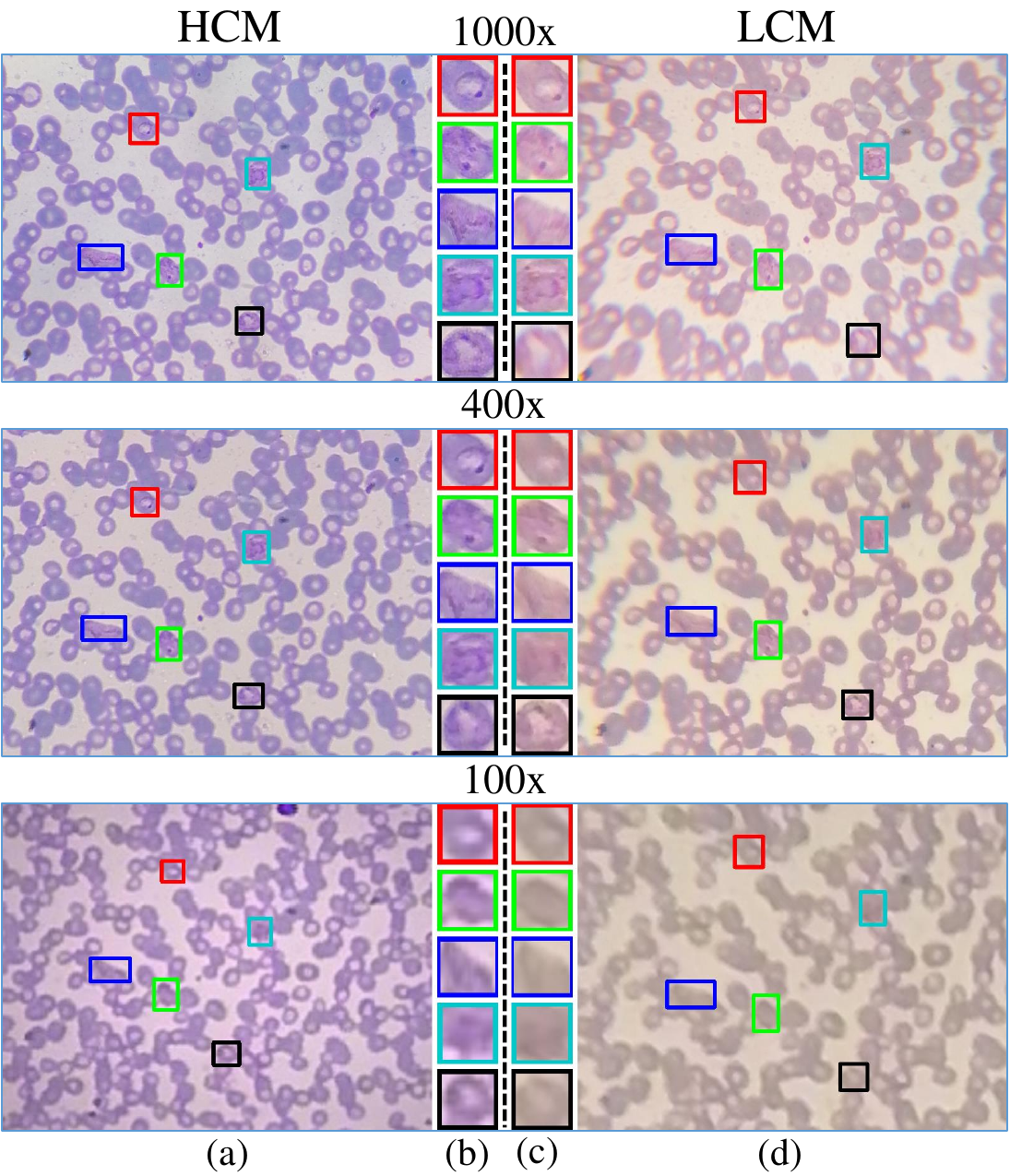}
    \caption{(a) shows images captured using high-cost microscope (HCM) at 1000\textsf{x}, 400\textsf{x} and 100\textsf{x} magnifications and (d) shows the same for low-cost microscope (LCM). (b) and (c) shows malarial cells from HCM and LCM respectively. Note that the images captured using HCM are clear and internal cells structures are more evident as compared to that of LCM. HCM cost 17 times more than LCM.
    } 
    \label{fig:LCM_HCM_Qual_comparison}
\end{figure}
Every year, on average, about 226 million cases of malaria are reported in 87 countries and 425,600 of them result in fatalities\footnote{average over the last five years} \cite{WHO}. Around 67\%  of all malaria deaths, in 2019, were of children under the age of five \cite{WHO}.
 Malaria, also named as a disease of the poor, mostly infects the population in resource deficient tropical and sub-tropical regions with a weak health care system.
 Although fatal, malaria is preventable and curable. 
 Even in developed countries, the leading cause of death in malaria is considered to be a delay in diagnosis and treatment \cite{CDC}.
Early detection not only allows avoiding the medical complications but also prevents further spread \cite{malariajournal}. 


Although the testing kits are becoming common, microscopic analysis of the stained blood slides is still considered a gold standard malaria diagnosis \cite{malariapaper,malariapaper2,Hung_2017_CVPR_Workshops, mehanian2017computer}. 
However, malaria microscopic analysis is a taxing process, requiring the availability of expensive microscopes and trained experts.
All these factors limit the accessibility to the early and correct malaria diagnosis, specifically in remote and resource constraint areas.
To tackle the subjectivity and shortage of doctors, several microscopic malarial image analysis methods have been proposed.  This includes approaches using hand-crafted visual features \cite{fatima2020automatic,linder2014malaria,molina2020sequential,tek2006malaria} and deep learning based approaches \cite{Hung_2017_CVPR_Workshops,vijayalakshmi2019deep,Ahmed_IEEEAccess,rajaraman2018pre,mehanian2017computer}. To achieve a good detection accuracy, these deep learning based algorithms require large datasets and their accuracy is dependent on the correctness of the annotation of these datasets. 
Although appealing, the previously proposed methods do not address the core limitations of malaria detection which is to detect malaria on low-cost microscopes (as malaria mostly occurs in resource-constraints areas) and on the low-magnifications lens (for efficient diagnostics). To achieve low-cost and efficient malaria detection using deep learning, it is vital to collect malaria data from the low-cost microscope and at lower magnifications.

\begin{figure}[t]
    \centering
    \includegraphics[width=\linewidth]{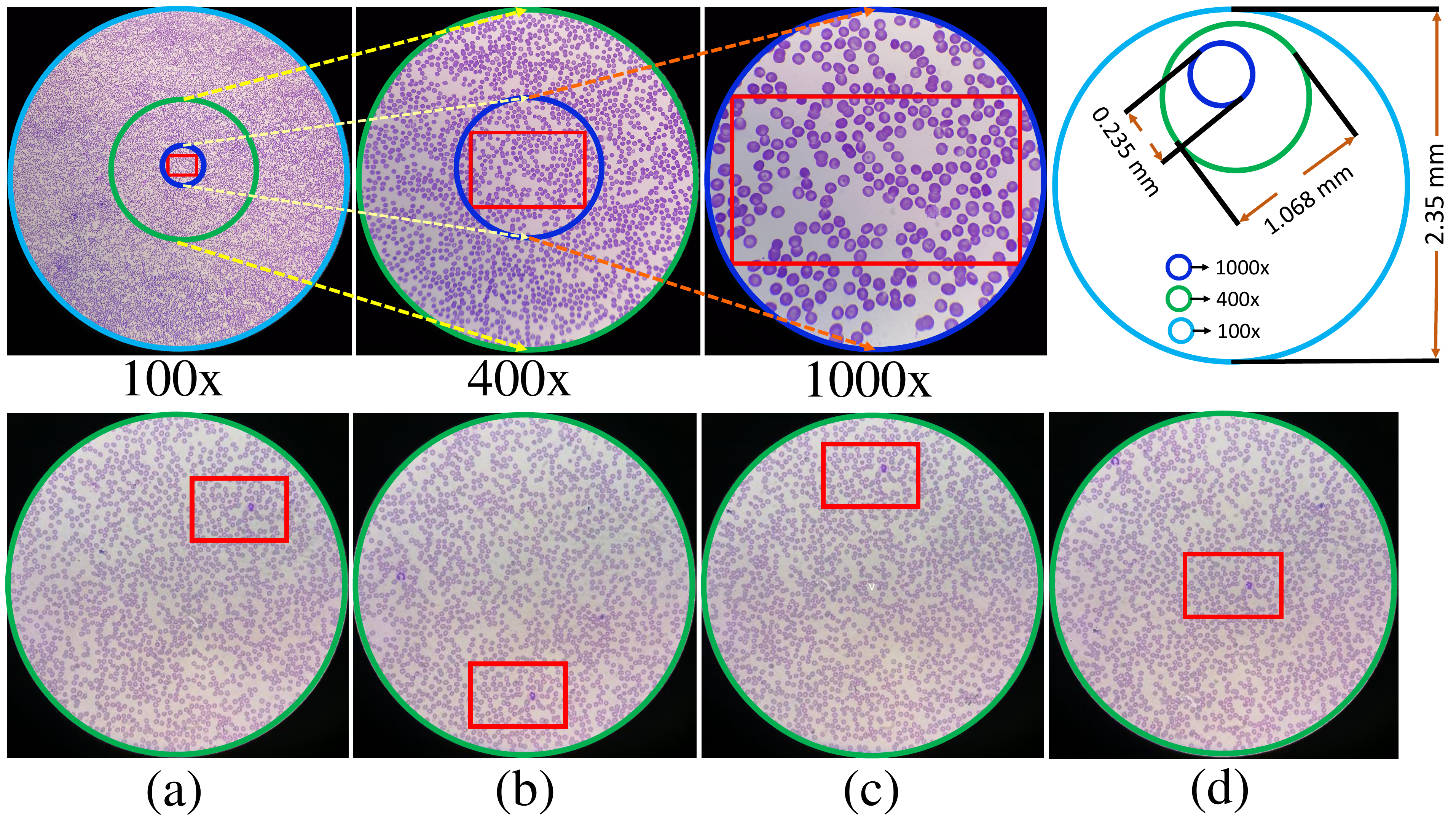}
    \caption{Top row presents the relationship between different magnification lenses of microscope. The second shows that the patch can appear at different locations in the same 400\textsf{x} FOV due to manual movement of microscope's stage. }
    \label{fig:mag_comparison}
\end{figure}

Low-cost microscopes (LCM), although more than 70\% cheaper than high-cost microscopes (HCM) suffer from a limited field of view (FOV) and less clear image due to the low-quality lens (see Figure     \ref{fig:LCM_HCM_Qual_comparison}). 
Due to these limitations, it is challenging and time-consuming for medical experts\footnote{Verified after contacting multiple doctors in different major hospitals.} to find the malaria cells at LCM and to confirm the malaria life-stage. 
On every microscope, the images could be captured at multiple magnifications such as 100\textsf{x}, 400\textsf{x}, and 1000{\textsf{x}}. Since 1000\textsf{x} makes the internal structure of cells very clear, it is a  standard practice by doctors to analyze malarial slides at 1000\textsf{x}.
FOV of 1000\textsf{x} is much smaller than that of 400\textsf{x} and 100\textsf{x} (see Figure \ref{fig:mag_comparison} and Section \ref{Preliminaries}). 
{As compared to 1000\textsf{x}, at 100\textsf{x} or 400\textsf{x} a much larger portion of blood-smear slide is visible at one time and therefore the whole slide could be traversed in less time for automatic malaria detection. Thus a higher accuracy at lower-magnification will be beneficial for efficient analysis.}



\begin{figure*}[t]
    \centering
    \includegraphics[width=0.90\linewidth]{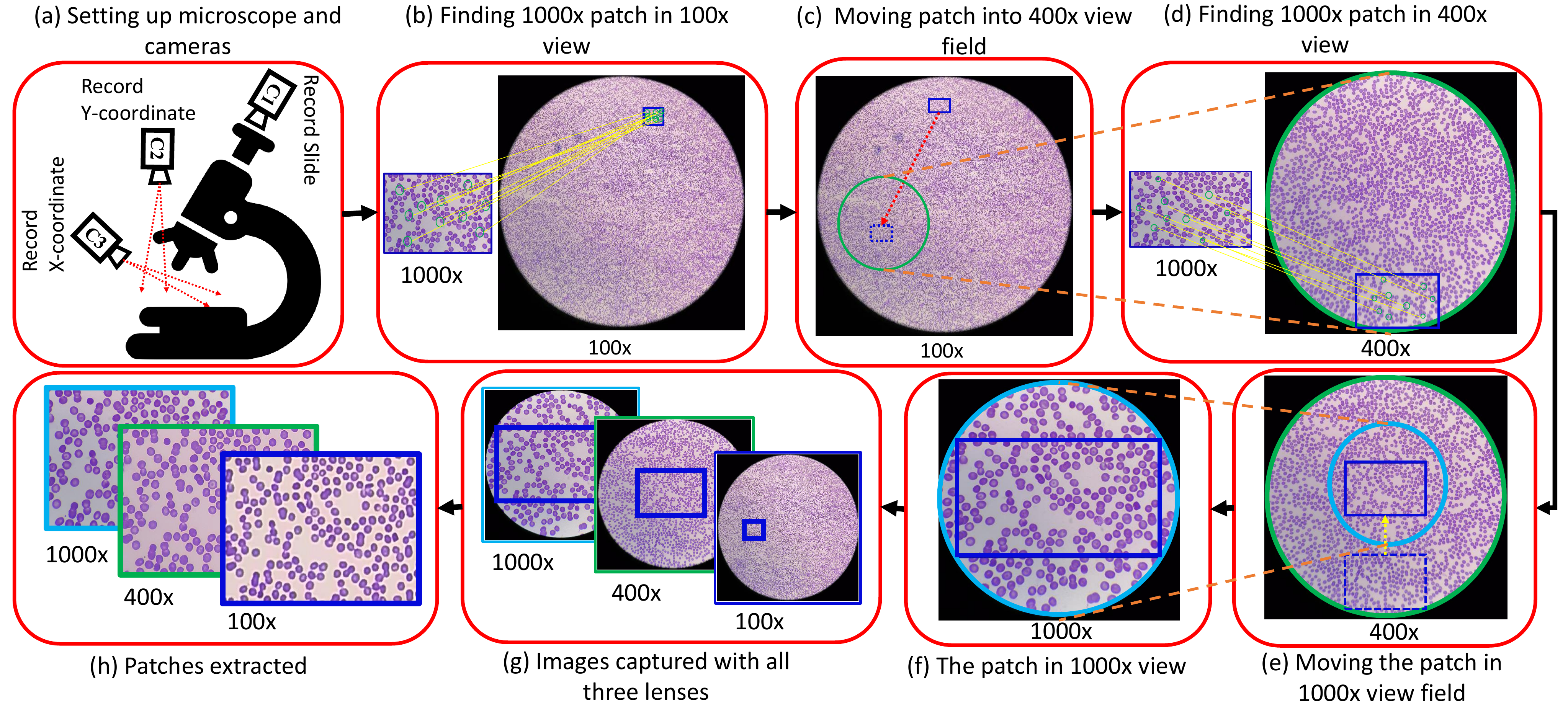}
    \caption{Steps to capture malarial region images at multiple magnification on HCM. (a) Initial image captured by hematologist at 1000\textsf{x} along microscope’s stage x-y locations pictures, (b) captured image is detected at 100\textsf{x} view, (c) 100\textsf{x} view image is tracked to a new location to be viewed at 400\textsf{x}, (d) The original 1000\textsf{x} patch is found in the 400\textsf{x} view followed by (e) tracking to new location to be found at 1000\textsf{x}, (f) finally the original 1000\textsf{x} patch is found at 1000\textsf{x}. (g)  and (h) shows the final extracted patches (See supplementary material).\vspace{-0.2cm}}
    
    \label{fig:dataset_collection}
\end{figure*}

To cater the above-mentioned challenges with low-cost microscopes, we put forward a new large scale \textit{multi-microscope multi-magnification malaria (\textbf{M5})} dataset. Our dataset contains images of the \textit{same} slide regions captured using HCM and LCM and at multiple magnifications (1000\textsf{x}, 400\textsf{x}, and 100\textsf{x}). As compared to the usual challenges faced in the data collection/annotations, computer-aided malaria data collection suffers additional challenges. First of all, the medical expert needs to manually locate all the malarial cells regions in the high-cost microscope at 1000\textsf{x} slides. 
Due to the mechanical parts and the properties of the lens, annotations collected in the high magnification are not easily transferable to low magnifications. A slight movement of the stage controls (due to manual manipulation) results in a large shift of the view, especially in high magnification. As we move from high magnification to low magnification, a portion of the slide in the center of the view (in high magnification) might not remain in the center of the view at the low magnification (see Figure \ref{fig:mag_comparison} and Figure  \ref{fig:dataset_collection}). 
Also, every time, we transverse the slide or change the magnifications, the slide needs refocusing.
Capturing the same HCM slide regions in LCM is further challenging.
As we move from the HCM to the LCM, FOV decreases resulting in loss of annotations, even if the center area is perfectly aligned (which itself is a challenging problem).
Therefore, to recover malarial cells regions \textit{manually} for collection of the dataset at multiple magnifications across multiple microscopes, we need to record and discover those locations manually at  \textbf{millimeter} accuracy. 

Our dataset can not only be used for the malarial cell localization and stage classification, but also for the across microscopes domain adaptation task. 
We provide baselines for both tasks on our dataset.
Multiple object detectors \cite{FASTERRCNN,FCOS,RetinaNet,YOLO} are trained and their results are reported for HCM at multiple magnifications. {As mentioned above, annotations for HCM are challenging to obtain but still comparatively easier than gathering annotations on LCM,} therefore, we show adaptation results for detectors trained on HCM (source domain) to the LCM (target domain). Finally, we also present our partially supervised domain adaptation strategy and report state-of-the-art results. 
In summary, this paper has following contributions:\newline\newline
  1) We collect the first large-scale multi-microscope multi-magnification malarial image dataset from the thin-blood smear slides. Dataset has been annotated by a medical expert with more than 20 years of experience, for both the malarial cell localization and life-stage classification tasks.\newline 
  2) An annotation mechanism (consisting of multiple mobile cameras and microscopes) is designed that records the microscope's x-y transnational control knobs positions to speed up the data collection, handles issues of calibrations across magnifications and also across microscopes. \newline
  3)  We compute the baseline results of several object detectors and domain adaptation methods on our dataset, creating a benchmark for the computer-aided malaria detection task.\newline 
  4) Finally,  utilizing the problem-specific constraints, a partially supervised domain adaptation mechanism is designed.

\section{Related work}\label{Related_work}

Recently, the vision community has shown a growing interest in collecting large-scale medical imaging datasets and bench-marking recent computer vision and deep learning models on them.  To make chest X-rays analysis low-cost and scalable, Phillips et al.,  \cite{pmlr-v136-phillips20a} presented a dataset containing pictures of chest X-rays captured by different mobile phone cameras.  To address commonly occurring thoracic diseases, Wang et al., presented chest X-rays dataset \cite{ChestXrays8} with image-level labels and provided weakly supervised baseline results. To facilitate Tuberculosis diagnosis, Liu et al., \cite{RethinkingTB} collected Tuberculosis X-ray (TBX11K) dataset containing X-ray images along with corresponding bounding box annotations. They provided experimental results of different object detectors on their dataset. 
Recently, domain adaptation strategies have been applied to medical imaging \cite{zia2021surgical, Surgvisdom, MIC2020DA}. 

Historically, taking a page from image analysis, several
malaria detection techniques were introduced using hand-crafted visual features. Authors in \cite{fatima2020automatic,di2002analysis,rao2002automatic,tek2006malaria, dave2017image} employed morphological operations to perform cell enhancement and segmentation followed by malarial cell classification using k-nearest neighbours, binary trees etc. 
Similarly, authors in \cite{kumarasamy2011robust,linder2014malaria,molina2020sequential} extracted different visual features such as SIFT \cite{SIFT2004}, local binary patterns \cite{LBP}, and classified cells using support vector machines \cite{SVM}.

Recently, several deep learning-based approaches have been presented for automatic malaria detection. Hung et al.,  \cite{Hung_2017_CVPR_Workshops} proposed a two-stage malaria detection approach. In their first stage, they applied Faster-RCNN to discriminate healthy versus malarial cells. During the second stage, they employed AlexNet \cite{AlexNet} for malarial stage classification. Vijayalakshmi et al., \cite{vijayalakshmi2019deep} proposed to extract CNN features from the VGG-16 network followed by SVM classification.
Umer et al., \cite{Ahmed_IEEEAccess} presented a stacked convolutional neural network for malaria detection in thin-film slides. Rajaraman et al., \cite{rajaraman2018pre} and Mehanian et al.,\cite{mehanian2017computer}
analyzed different  CNN architectures for malaria detection on thin blood smears and think blood smears respectively. Both methods used traditional thresholding-based approaches for cell segmentation. Finally, some malaria datasets have been proposed recently \cite{BBBC041,boray2010,loddo2018mp,arshad2021dataset}.

As compared to the above-mentioned approaches, our main focus in this paper is to collect a multi-microscope multi-magnification malaria dataset and evaluate different CNN object detectors and domain adaptation methods.

\label{sec:dataset}

\section{Preliminaries}\label{Preliminaries}
\noindent\textbf{Malaria:}  Due to low cost, high sensitivity, and specificity, an optical microscope is a gold standard for diagnosing malaria by examining thin blood smears \cite{world2019world}. This is in contrast to Polymerase Chain Reaction
(PCR) \cite{fatima2020automatic} and Rapid Diagnostic Test (RDT) \cite{diaz2009semi}, which are either expensive or have low accuracy.
As compared to other malaria types,  malarial parasites named Plasmodium vivax (P. vivax) are more common and fatal, therefore we collected data from P.vivax infected patients. P.vivax has four stages (classes) of development in the human body. These stages are ring, trophozoite, schizont, and gametocyte. Classification of a cell infected by malaria into these four classes is crucial to identify the seriousness of the disease and for the prescription of the best medication. \newline
\noindent\textbf{Relationship between different microscopes:}  The quality and price of microscopes are governed by their field of view, precise movement of the mechanical stage, and most importantly the quality of the magnification lenses (e.g., plan achromatic versus non-plan achromatic). The price range for good quality microscopes is around several thousand dollars.  In our setup, the price of the high-cost microscope (Olympus CX23) is around 3,000 USD
 and the price of the low-cost microscope (XSZ-107BN) is around 160 USD.
The high-cost microscopes, although preferred by hematologists and lab technicians, are too expensive to be readily available at all locations. 
At the same time, resource-constrained areas have very few highly trained experts available. 
These two constraints result in the situation where automatic methods for the microscopic analysis of the slides, especially for the low-cost microscopes are highly needed.

\noindent\textbf{Relationship between different microscopic magnifications:} The magnification of a microscope is measured by eyepiece magnification $\times$ objective lens magnification. Most of optical microscopes are equipped with three magnification scales: 100\textsf{x}, 400\textsf{x} and 1000\textsf{x} which makes cell/tissues look 100 times, 400 times, and 1000 times bigger respectively. Note that 1000\textsf{x} is also called immersion oil magnification since oil must be placed on blood smear to see clear cell structure at 1000\textsf{x}. %
In Figure \ref{fig:mag_comparison} (top row), we depict the relationship between different magnifications.A FOV on 100\textsf{x} covers around 20 FOVs of 400\textsf{x} and almost 180 FOVs of 1000\textsf{x}.  For clarity purposes, we manually moved the patch of interest to the center location. In reality, the \textit{same} patch appear may appear in different locations at different magnifications.
The patch correspondence between magnifications is different for different microscopes and is discovered during the microscope calibration step.
As shown in the second row of Figure \ref{fig:mag_comparison}, due to the manual setting (in millimeter)  of the microscope's stage, on repetition, the same patch can appear at different locations.



 
\section{Dataset Collection Process}


In the process of understanding the problem and data collection, we contacted multiple health professionals and multiple hospitals. 
Collecting the slides of malarial infected patients and annotating them, was a challenge. 
During this process, it was discovered that, due to the low quality of visual information from the LCM, experts were reluctant to use LCM for diagnosis or annotation. 
Therefore we decided to only annotate data using HCM at 1000\textsf{x}.
To tackle the above-mentioned challenges, we propose to capture images of malarial regions and annotations on HCM at 1000\textsf{x} and then transfer these annotations to multiple magnifications in and across microscopes. 

Dataset collection consists of multiple steps and multiple people. 
  Collecting the blood-smear slides from the malarial-infected patients is itself a challenging process.
  First of all, images of malarial regions are captured by the expert hematologist, by traversing the slide on HCM at 1000\textsf{x}, along with images of the exact microscope’s x-y transnational control knobs coordinates. 
  In addition, hematologist provides helps localize (dropping a pin on) the malaria-infected cells and identify their life-stage labels. 
  The operator then uses this information to draw the bounding boxes.
  After that, we locate these malarial regions at multiple magnifications on both HCM and LCM. 
  These localizations are manually verified.
The complete dataset is
finalized after intense efforts of several months. All personal information of patients has been removed to make the dataset publicly available. Table~\ref{table:dataset_comparison} demonstrates the comparison of our dataset with existing malaria datasets. None of
the existing datasets capture malaria from different microscopes
or at multiple magnifications. 

  

\vspace{-0.3cm}
 \subsection{Initial Malarial Image Collection}
 \label{InitialMalarialImageCollection}


To annotate and collect images, we seek help from a hematologist having more than 20 years of experience.
An annotation rig is designed, consisting of three mobile phone cameras and microscopes. 
Expert traverses the slides to identify both the images with the malarial cells and localize and classify these cells.
The mobile camera is mounted on a microscope eyepiece using a cell phone stand (C1 in Fig. \ref{fig:dataset_collection} (a)), to capture the image. 
Since we are aiming at a low-cost solution, we use a nominal price ($<$200 USD) mobile phone camera (Honor 9x Lite) to capture images without any post-processing.
To transfer these annotations from one magnification to other and across the microscope, two cameras are set up (C2 \& C3 in Fig. \ref{fig:dataset_collection} (a)) around the microscope to record the microscope's stage x-y locations. We need two cameras to avoid occlusion from microscopic nose-piece objective lenses.
This allows the more efficient annotation and helps save experts time. 
Note that each image is captured and each malarial cell is localized by the hematologist.

\subsection{Image Capturing  on HCM}

Our next task is to collect malarial images on different microscopic magnification scales on HCM.
As shown in Figure \ref{fig:dataset_collection} (b), after extracting the maximum size rectangular patch that covers most of the cells at 1000\textsf{x}, we search that patch on 100\textsf{x}  FOV. 
100\textsf{x} FOV covers almost 180 FOV of 1000\textsf{x}, as discussed in the Section~\ref{Preliminaries} under \textit{Relationship between different microscopic  magnifications}.
We do not search the patch on 400\textsf{x} directly from 1000\textsf{x} as the small change in the stage location (due to the manual movement) drastically changes the view and makes it almost impossible to find the exact malarial patch. 
To search the patch in 100\textsf{x} FOV, we employ keypoint detection (Speeded-Up Robust Features (SURF) \cite{SURF}) followed by RANSAC \cite{RANSAC}. 


After cropping the patch in the 100\textsf{x}, we want to find the patch in the 400\textsf{x}. 
However, the center visible in 100\textsf{x} magnification does not map to the center of the view when magnification is increased while keeping the x-y location fixed (Figure~\ref{fig:dataset_collection} (c)). 
We perform cross magnification calibration to identify where the region from the lower magnification is shifted in higher magnification. 
A graphical interface is designed to guide the operator to traverse the slide manually such that the detected patch could move to the location that can be seen after switching to a 400\textsf{x} magnification (look at the green circle in Figure~\ref{fig:dataset_collection} (c)).
Note that this step is accomplished by simultaneous manual movement of the slide and real-time tracking of the patch.
Global image alignment is performed to track the patch across the frames as the operator transverse the slide. Once the microscope magnifications is changed to 400\textsf{x}, 1000\textsf{x} patch is searched again in the new FOV (Figure~\ref{fig:dataset_collection}(d)). We need this search since the patch may exist anywhere within 400\textsf{x} FOV due to manual movement of the microscope's stage by the operator.

Above steps are repeated as we move from 400\textsf{x} to 1000\textsf{x} magnification\footnote{We search 1000\textsf{x} again to have a good quality picture of 1000\textsf{x} which is sometime difficult in front of doctor due to time constraints.} (Figure~\ref{fig:dataset_collection} (e), and (f)).
Figure~\ref{fig:dataset_collection} (g) and (h) shows the final patch extracted at 100\textsf{x}, 400\textsf{x} and 1000\textsf{x}. 
Note that we need to apply immersion oil at 1000\textsf{x} and rinse it off at 100\textsf{x} and 400\textsf{x} from the slide for every single image. 
The procedure is repeated for all malarial regions. 

\subsection{Image Capturing on LCM}
After capturing images at multiple magnifications on HCM, our next task is to capture the same malarial regions on LCM. Since we already have recorded stage x-y coordinates of malarial regions of the slide on HCM, we manually set the LCM stage to those coordinates. 
This step requires across microscope \textit{stage calibrations} as stage coordinate for both microscopes for the same slide-location are shifted.

The overall procedure for finding the malarial patches on LCM is similar to that of HCM except for a few changes which we discuss below (see supplementary material).  
At step (b), (d) and (f) shown in Figure \ref{fig:dataset_collection}, instead of patch from 1000\textsf{x} magnification, 100\textsf{x}, 400\textsf{x}, and 1000\textsf{x} magnification patches from HCM are searched at 100\textsf{x}, 400\textsf{x}, and 1000\textsf{x} FOV of LCM. The rest of the steps (c), (e) remain the same.


\begin{figure}[t]
    \centering
    \includegraphics[width=\linewidth, height=3.1in]{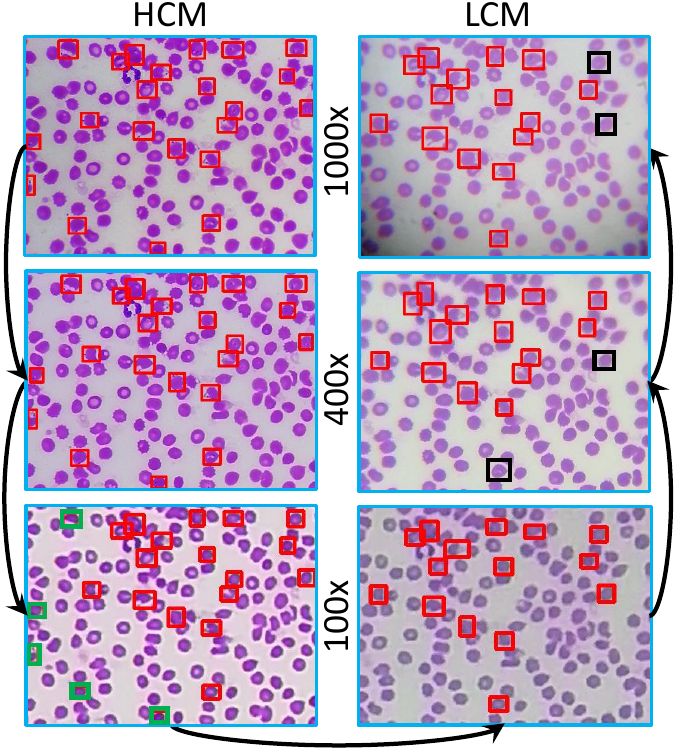}
    \caption{Transferring annotations across magnifications and across HCM and LCM. The green boxes shows the malarial cells that are not in the LCM FOV (as LCM has smaller FOV than HCM). The black box's annotations were missed during automatic transformations and were included during manual verification.}
    \label{fig:Annotation_Transfer}
\end{figure}
\subsection{Malarial Cell Annotations on HCM and LCM}
For any given malarial region annotated by an expert hematologist with bounding box and class labels at 1000\textsf{x} on HCM, these annotations are automatically transferred to multiple magnifications across the microscope. 
To make this transfer efficient, we first compute homography between the images using SURF feature point matching along with RANSAC. Figure \ref{fig:Annotation_Transfer} summarizes the annotation transfer process.
Specifically, on HCM,  we transfer the bounding box annotations from 1000\textsf{x} to 400\textsf{x} and  from 400\textsf{x} to 100\textsf{x}. Annotations are transferred from 100\textsf{x} image of HCM to 100\textsf{x} image of LCM. Finally, on LCM, the annotations are transferred from 100\textsf{x} to 400\textsf{x} and from 400\textsf{x} to 1000\textsf{x}. Note that due to the difference in resolutions, image quality, and FOVs, we often get poor key points matching or insufficient features points to be matched to compute homography. Therefore, after automatic annotation transfer, we manually verified each image to correct shifted bounding boxes and put missing bounding boxes.

\begin{table}
\footnotesize
\begin{tabular}{|c|c|c|c|c|c|}
\hline
Malaria Dataset & \multicolumn{1}{c|}{\begin{tabular}[c]{@{}c@{}} Across\\Micros.\end{tabular}} & \multicolumn{1}{c|}{\begin{tabular}[c]{@{}c@{}} Multi\\Magn.\end{tabular}}  &  \multicolumn{1}{c|}{\begin{tabular}[c]{@{}c@{}} Malarial\\cells \end{tabular}}& 
\multicolumn{1}{c|}{\begin{tabular}[c]{@{}c@{}}BBX  \end{tabular}} &  
 \multicolumn{1}{c|}{\begin{tabular}[c]{@{}c@{}} No. of\\Images\end{tabular}}
  \\ \hline
\hline
BBBC041\cite{BBBC041} & No & No &   {$2452$} & Yes & 1364\\
Malaria655\cite{boray2010} & No & No &   {557} & No & 655 \\
MPIDB\cite{loddo2018mp} & No & No &   {840} & No & 229 \\
IML\cite{arshad2021dataset}   & No & No &  {$529$}  & Yes & 345 \\
\hline
\hline
Our dataset & \textbf{Yes} & \textbf{Yes} &  \textbf{20,331}* & \textbf{Yes} & \textbf{7542}*\\ 
\hline

\end{tabular}
\caption{Comparison of malaria datasets. Medical expert traversed multiple slides identifying 1257  malarial regions using HCM at 1000\textsf{x}. Images corresponding to these locations were collected at all the magnifications in both HCM and LCM microscopes resulting in the dataset of size 7542 ($=2\times 3 \times 1257$) and 
total number of malarial cells 3$\times$3624 (HCM) + 3 $\times$ 3153 (LCM) = 20331.}
\label{table:dataset_comparison}
\end{table}



\section{Evaluation Methods} 
We provide an initial benchmark by evaluating two types of methods on our dataset. 
To compute the accuracy of malarial cells detection as well as malarial-cell stage classification, we apply recent object detectors. 
Our experimental results (Table~\ref{tab: ObjectDetecorsResult_Across_mags}) show that detector trained on HCM performs poorly on LCM even on the same magnification. Therefore, we evaluate a few recent domain adaptation approaches to improve detection accuracy across the microscopes. 
Furthermore, we propose to use problem-specific constraints to improve cross microscope detection accuracy. 
Our key findings are that these problem-specific constraints can be plugged-in into any domain adaptation approach and can help improve cross-microscopes accuracy.

\subsection{Object Detectors}
To generate benchmark results on our dataset, we perform experiments using recent popular object detectors including  Faster R-CNN \cite{FASTERRCNN}, YOLO-v3 \cite{YOLO}, RetinaNet \cite{RetinaNet}, and FCOS \cite{FCOS}.
YOLO\cite{YOLO}, is a single-stage detector that divides the image into grids to achieve a fast inference rate. 
RetinaNet \cite{RetinaNet} employs novel focal loss to cater to the class imbalance. 
Instead of relying on a set of pre-defined anchor boxes to compute proposals, FCOS \cite{FCOS} uses the center point of objects to define whether a location is positive and regresses the four distances from the center to the object boundary. 
YOLO, RetinaNet, and FCOS are single-stage whereas Faster R-CNN is a two-stage detector.

\subsection{Domain adaptation Methods}
\label{sec:DA}
Our benchmark object detectors report reasonable results over the HCM (Table~\ref{tab: ObjectDetecorsResult_Across_mags}), however, their accuracy drops when tested on LCM. 
Due to several factors including the lens differences, images taken from the  LCM exhibit considerable domain shift from the ones taken from HCM. In 
Figure~\ref{fig:LCM_HCM_Qual_comparison}, we present some of the sample images representing distribution shift. This shift is exhibited not only in terms of change in color distribution but also that LCM images appeared to be more blurred and less clear, with the fine-grain texture information reduced when compared to the HCM images. We explore different domain adaptation strategies to overcome domain shift. 

Since Faster-RCNN has the highest mAP in comparison to other detectors (Table~\ref{tab: ObjectDetecorsResult_Across_mags}),  we evaluate the object detector domain adaptation strategies \cite{xu2020cross, saito2019strong, chen2018domain} that are based on Faster RCNN on our dataset. 
 Xu et al., \cite{xu2020cross} propose Graph-induced prototype alignment for the domain adaptation and report results on cityscapes to foggy-cityscapes datasets, a case similar to ours. Chen et al., 
 \cite{chen2018domain} controls the image and instance-level shifts to avoid domain discrepancy at an optimal level whereas \cite{saito2019strong} suggests strongly aligning the local features and weakly aligning the global features to retain the unique features of each domain that may help in better inferences. 
 However, all these methods are unable to tackle the domain shift in our case, hence we propose a domain adaptation method specific to our dataset. 

\begin{figure}[t]
    \centering
    \includegraphics[width=\linewidth]{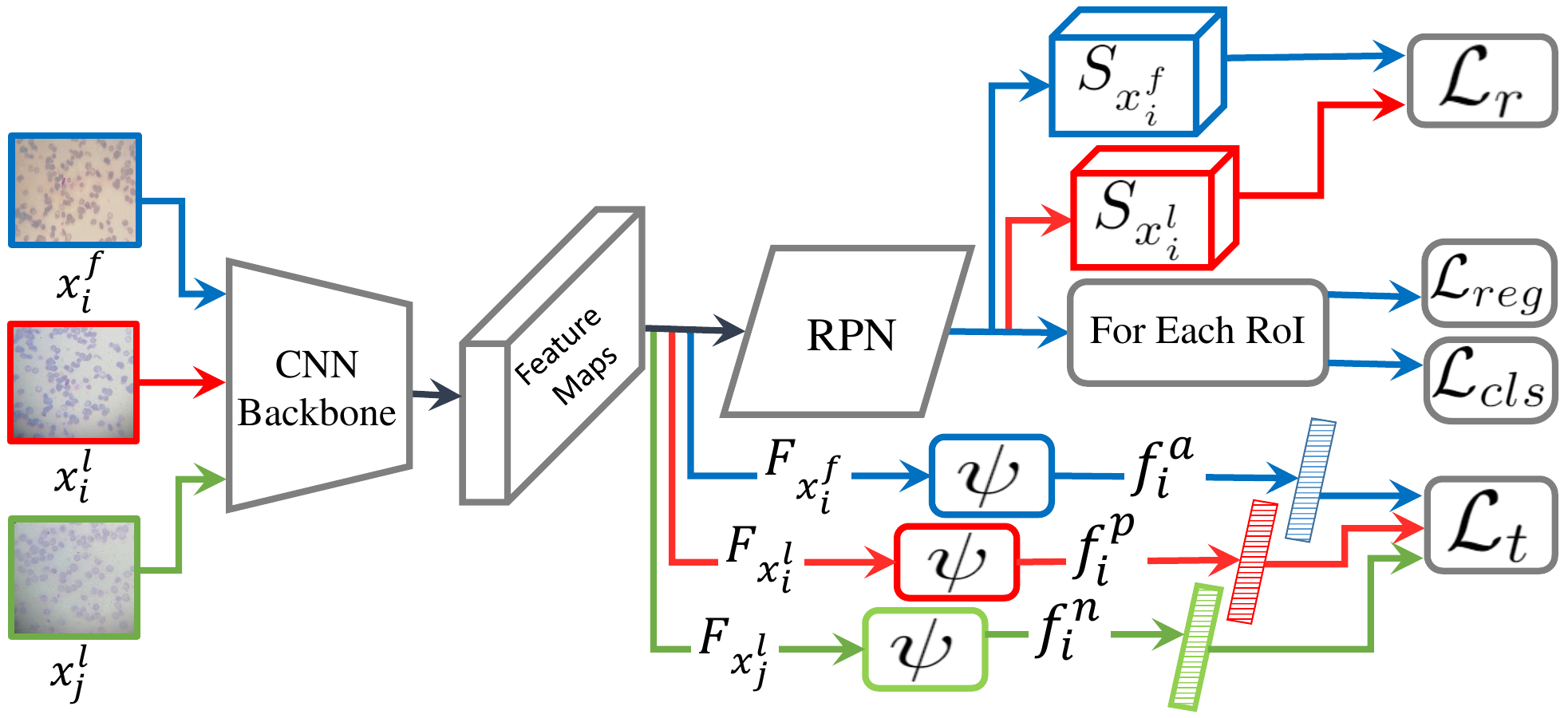}
    \caption{ Partially supervised domain adaptation method. For details, see Section~\ref{sec:DA} }
    \label{fig:implementation}
\end{figure}

\begin{table*}[t]
\footnotesize
\centering
\begin{tabular}{|c?c|c|c?c|c|c?c|c|c?c|c|c?}
\hline
{\begin{tabular}[c]{@{}c@{}}Training\\Magnification\end{tabular}} &  \multicolumn{3}{|c|}{FCOS}& \multicolumn{3}{c|}{RetinaNet} & \multicolumn{3}{c|}{YOLO}&\multicolumn{3}{c|}{Faster R-CNN}  \\ \cline{2-13} 
                              & \multicolumn{12}{|c|}{Test Magnification} \\ \cline{2-13} 
                             & \multicolumn{1}{|c|}{1000\textsf{x}} & 400\textsf{x} & 100\textsf{x} & \multicolumn{1}{|c|}{1000\textsf{x}} & 400\textsf{x} &100\textsf{x} &\multicolumn{1}{|c|}{1000\textsf{x}}& 400\textsf{x} & 100\textsf{x} & \multicolumn{1}{|c|}{1000\textsf{x}} & 400\textsf{x} & 100\textsf{x} \\ \hline
                   \hline
1000\textsf{x} & \multicolumn{1}{|c|}{36.8} & 13.5 & 0.0& \multicolumn{1}{|c|}{43.1} & 29.7 & 0.0 & \multicolumn{1}{|c|}{62.8}& 36.7 & 0.0 & \multicolumn{1}{|c|}{\textbf{66.8}} & 31.3 & 0.0  \\ \hline

 400\textsf{x} & \multicolumn{1}{|c|}{31.4} & 29.1 & 1.9 & \multicolumn{1}{|c|}{32.9} & 34.0 & 1.8 & \multicolumn{1}{|c|}{55.2} & 56.6 & 4.5  & \multicolumn{1}{|c|}{56.9} &\textbf{61.1} & 1.4  \\ \hline 
 
 100\textsf{x} & \multicolumn{1}{|c|}{9.4} & 14.8 & 8.9 & \multicolumn{1}{|c|}{10.2} & 15.4 & 16.3 & \multicolumn{1}{|c|}{10.5} & 3.9 & 20.1& \multicolumn{1}{|c|}{25.4} & \textbf{31.9} & 31.5 \\ \hline
 
\end{tabular}
\vspace{-0.1cm}
\caption{Experimental results of models trained and tested on different magnifications on HCM. The evaluation metric is mAP.}
\label{tab: ObjectDetecorsResult_Across_mags}
\end{table*}

\noindent\textbf{Partially Supervised Domain adaptation Approach:}
  Let $ D_h = \{(x_i^h, y_i^h)\}_{i=1}^{N_h}$ be the set of images obtained from the HCM along with their annotations, and $ D_l = \{(x_i^l)\}_{i=1}^{N_l}$ be the images obtained from LCM. 
 We assume that both $D_h$ (source domain) and $D_l$ (target domain) have same magnification and $N_h=N_l$.
 Note that we do not have the annotations for the $D_l$, however, we do have the correspondence across the datasets. 
 That is $x_i^h \in D_h$ and $x_i^l \in D_l$ are obtained by centering microscopes to the same region of the slide. 
 However, due to the decrease in FOV in the LCM, regions captured in both microscopes might not be exactly the same. 
 In fact, $x_i^l$ might have slightly less number of cells than the $x_i^h$. 
 We use this correspondence information to come up with the \textit{partially-supervised domain adaptation} strategy. 
 
 First, using the CycleGAN\cite{CycleGAN2017} $\mathit{G_{h\rightarrow l}}$, we translate the images $x_i^h \in D_h$ to $D_l$. 
 Thus creating an intermediate fake-LCM domain $D_f=\{(x_i^f = \mathit{G_{h\rightarrow l}}(x_i^h), y_i^f=y_i^h)\}_{i=1}^{N_h}$.  
Different strategies are used to perform domain alignment at different stages of  Faster-RCNN.
 Since we have access to the ground-truth for the intermediate fake-LCM domain $D_f$, we can use the standard loss functions ($\mathcal{L}_{reg}$ and $\mathcal{L}_{cls}$) to bridge the gap between the  $D_l$ and $D_h$. 
We introduce ranking loss and triplet loss to align the feature extraction and the region proposal network (RPN) across domains. 

Let $F_x$ be the feature volume being input to the RPN and $S_x  \in \mathbb{R}^{(H\times W \times K)}$ be the objectness score output by the RPN for $K$ anchors when we input image $x$.
Since, $x_i^f = \mathit{G_{h\rightarrow l}}(x_i^h)$ and $x_i^l$ are corresponding images across the domain, RPN should have a similar response for both. 
In order to minimize the difference between objectness score of the proposed region from the RPN for fake-LCM and its corresponding LCM image, we use the following ranking loss: 
\begin{equation}
\vspace{-0.2cm}
\mathcal{L}_r(x_i^f, x_i^l) = max(0,   avg(S_{x_i^f}) - avg(S_{x_i^l})-\beta),
\label{eq:rankingLoss}
\end{equation}where $\beta$ is the margin.
Similarly based on the correspondence across domains, we introduce the \textit{triplet loss} to maximize the similarity between $x_i^f$ and $x_i^l$. 
Let $x_i^f$,  $x_i^l$ and $x_j^l$ ($i \ne j$) be anchor, positive and negative samples for the triplet loss. 
Let $f_i^a= \psi(F_{x_i^f})$, $f_i^p= \psi(F_{x_i^l})$ and $f_i^n= \psi(F_{x_j^l})$, where $\psi$ is  global average pooling function, the triplet loss is defined as: 
\begin{equation}
 \mathcal{L}_t = max(0, ||f_i^p - f_i^a||_2^2 -  ||f_i^a - f_i^n||_2^2 + \alpha),
 \end{equation}
where $\alpha$ is the margin. 
The source domain model is adapted using the standard Faster-RCNN loss-function for the $x_i^f \in D_f$ (images from the intermediate domain). 
For the images from the $D_l$,  the $\mathcal{L}_t$ and  $\mathcal{L}_r$ help in domain alignment. 


 


\section{Experimental Results}\label{ExperimentalResults}
\subsection{Implementation details}

All the object detectors are trained using \textit{mmdetection library} \cite{mmdetection}. 
 ResNet-50\cite{ResNet} (pre-trained on Imagenet) with feature pyramid network is used as backbone except for the YOLO\cite{YOLO} which uses the Darknet53\cite{YOLO}. 
 We use stochastic gradient descent with a base learning rate of $0.01$ and momentum of $0.9$ for all detectors except for the FCOS\cite{FCOS} where we set the base learning rate to $0.001$.
 All models are trained for 30 epochs except 100 epochs for the YOLO\cite{YOLO} and the learning rate is divided by 10 after the eighth and twelfth epoch.
Faster-RCNN based methods domain adaptation  \cite{xu2020cross, saito2019strong, chen2018domain} are evaluated using the author's provided codes.  $\beta$ and $\alpha$  are chosen 0 and 1, respectively.


\noindent\textbf{Dataset splits:}
Our train set contains 66.5\% of the total images and also around 66\% of each malaria class; our test set contains 30\% of the total images and from 27\% to 30\% of each malaria class and finally, our validation set contains 3.5\% of the total images and from 3.5\% to 5.5\% of each malarial class. We have same division across all the magnifications (i.e., 100\textsf{x}, 400\textsf{x} and 1000\textsf{x}) and across the microscope (HCM and LCM), 
with corresponding images being in the same split across magnifications and microscopes.


\begin{figure*}[tb]
    \centering
    \includegraphics[width=0.98\linewidth]{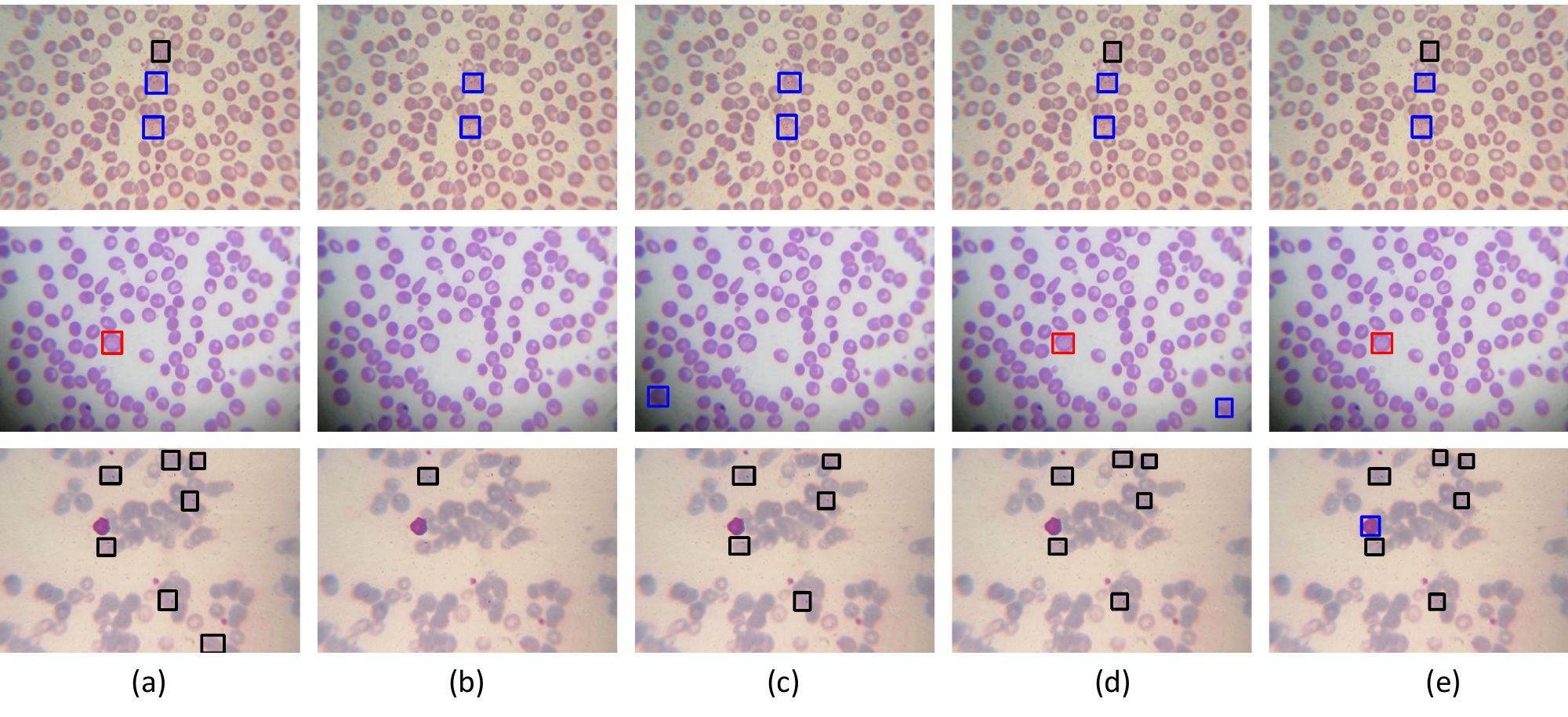}
    \vspace{-0.5cm}
    \caption{ 
Each row represents different malarial regions of LCM 1000\textsf{x}.
    {\textcolor{red}{$\square$}}, \textcolor{black}{$\square$}, \textcolor{blue}{$\square$} represent {gametocyte, ring, and trophozoite} respectively. (a) ground truths, (b) results of source only model, (c) results for HCM trained model, fine-tuned on fake LCM with ranking loss. (d) shows results when model trained on HCM, fine-tuned on fake LCM with triplet loss, and (e) shows results with both Ranking \& Triplet loss. 
    }
    \label{fig:q.r_success}
\end{figure*}

\subsection{Experimental results of Object Detectors}\label{ObjectDetectorResults}


Table \ref{tab: ObjectDetecorsResult_Across_mags} shows results of above mentioned detectors when trained separately for all three magnification levels on HCM.
Faster R-CNN  exhibits overall better mAP (mean average precision). 
In almost all the experiments, the best results are obtained when both training and testing are done on the same magnification level. 
For all detectors, the detection accuracy is quite low at 100\textsf{x} even when the model is trained on 100\textsf{x}. We believe that this is because even on HCM,  the internal structure of cells (which is important for malarial cell detection and stage classification) is not clearly visible at 100\textsf{x} (Figure \ref{fig:LCM_HCM_Qual_comparison}). 
Due to low accuracy at 100\textsf{x}, we decided to focus only on 1000\textsf{x} and 400\textsf{x} for later experiments. 
{Where models trained on 400\textsf{x} magnification give reasonable results on the 1000\textsf{x} (even comparable to models trained on 1000\textsf{x}), vice versa is not true. This could be attributed to the fact that supervised deep learning models are infamous for learning shortcuts and over-fitting the domain. Whereas noise and less fine-grained texture information in 400\textsf{x} images allow the model to be more robust.}
Since the FOV at 400\textsf{x} covers around 20 FOVs of 1000\textsf{x}, we can scan the blood smear slide and perform malarial cell detection about 20 times faster at 400\textsf{x} as compared to scanning at 1000\textsf{x}. 

\begin{table}[b]
\footnotesize
    \centering
    \begin{tabular}{|c|c|c|}
    \hline
    {\begin{tabular}[c]{@{}c@{}}Methods\end{tabular}} & \multicolumn{2}{|c|}{Source = HCM, Target = LCM}\\\cline{2-3} 
                              & 1000\textsf{x} $\rightarrow{}$ 1000\textsf{x} & 400\textsf{x} $\rightarrow{}$ 400\textsf{x} \\ \hline
    Xu et al.\cite{xu2020cross} & 15.5 &\textbf{21.6}\\
        Saito et al.\cite{saito2019strong} & \textbf{24.8} & 21.4\\
        Chen et al.\cite{chen2018domain}& 17.6 & 21.5\\ \hline
    Source only& 17.1 & 26.7 \\
    Fine Tuning on fake-LQM & 33.3& 31.8  \\
        Ranking loss & 35.7 & 32.4 \\
        Triplet loss & 37.2 & 32.2 \\ 
        Ranking+Triplet loss  & \textbf{37.5} & \textbf{33.8}\\\hline
    
    \end{tabular}
    \vspace{-0.2cm}
    \caption{Results (mAP \%) of domain  adaptation methods.}
    \label{tab:DA_comparison}
\end{table}

\subsection{Experimental results of Domain adaptations}
The experimental results of adapting from one magnification in HCM (source) to a similar magnification in the LCM (target) are detailed in Table \ref{tab:DA_comparison}. 
The method proposed in  \cite{saito2019strong}, outperforms the other two off-the-shelf algorithms.
One reason could be that it has explicit local alignment and malaria cell localization is less dependent upon the context.  However, results are still lacking.  

The experimental results with \textit{ranking loss} and \textit{triplet loss} show a significant improvement as compared to other methods. Triplet loss works better on 1000\textsf{x} as compared to ranking loss and ranking works better at 400\textsf{x} as compared to triplet loss. Both losses have a complementary effect as their combination improves the overall results. 
For comparison, we present results of FasterRCNN trained on HCM (source only) and also model fine-tuned on the fake-LCM (since we have ground truth for that). Note that we do not have ground truth for the training part of LCM. 
We evaluated our method for the across magnifications adaptation. For 
For FasterRCNN, HCM (1000\textsf{x}) to LCM (400\textsf{x}) with our approach mAP improves from 1.4 to 13.2. 
Adaptation was not successful for HCM (400\textsf{x}) to LCM (100\textsf{x}). 

\subsection{Qualitative Analysis}
Some of qualitative examples when model is trained on HCM (1000\textsf{x}) and tested on LCM (1000\textsf{x}) with and without domain adaptation are shown in Figure \ref{fig:q.r_success}.
The first row shows that source only model in (b) successfully detects trophozoite but is unable to detect the ring. Equipped with triplet loss, our model is able to predict the missing ring successfully.
The second row shows that although 
ranking and triplet loss give a false positive at different locations, their combination results in false-positive removal, on the other hand, the source-only model does not provide any detection. 
Finally, in the third row, we showed a failure case. 
As compared to the source model, our model is able to detect several rings, however, it misses the ring at the bottom. 
Also,  neutrophil has been miss-classified into trophozoite due to its appearance similarities. We consider these results quite encouraging since adapting from the image in the clear domain to the ones with less fine-grain information is a challenging task. 

\section{Conclusion}\label{conclusion}
We have attempted making malaria detection low-cost and efficient. 
To this end, we have collected a large-scale multi-magnification, multi-microscope malaria dataset. Furthermore, we benchmark different popular object detectors and domain adaptation methods on our dataset. We believe that this dataset will pave the way for future research in developing algorithms that can work with low-cost microscopes. The malaria-cycle stage contains inherent class imbalance because most blood samples are usually gathered at the middle of the malaria life cycle when patients show  symptoms. Our future work will be to focus on designing detectors that handle class imbalance.

\noindent\textbf{Acknowledgement:} We thank Qazi Ammar Arshad and Muhammad Affan for their help and contribution in the project. 
The project is partially supported
by an unrestricted gift award from Facebook, USA. 




{\small
\bibliographystyle{ieee_fullname}
\bibliography{egbib}
}

\end{document}